\documentclass{article}
\usepackage{iclr2021_conference,times}

\usepackage[utf8]{inputenc}
\usepackage[english]{babel}
\usepackage{url}
\usepackage{mathtools}
\usepackage{amsfonts}
\usepackage{hyperref}
\usepackage{epigraph}
\usepackage{vhistory}
\usepackage{grffile}

\usepackage{xcolor}
\usepackage{amsmath}
\usepackage{graphicx}
\usepackage{capt-of}
\usepackage{booktabs}
\usepackage{varwidth}
\usepackage{subfig}

\usepackage{amsthm}
\theoremstyle{definition}

\title{Challenges and approaches to privacy preserving post-click conversion prediction}

\author{Conor O'Brien\thanks{These authors contributed equally.}  \:, Arvind Thiagarajan$^*$, Sourav Das, Rafael Barreto, \\ \textbf{Chetan Verma, Tim Hsu, James Neufield, Jonathan J Hunt}
\\
\texttt{\{conoro, arvindt, sdas, rbarreto,}
\\ \: \texttt{cverma, timh, jneufeld, jjh\}@twitter.com}
}

\iclrfinalcopy

\begin{document}

\maketitle

\begin{abstract}
Online advertising has typically been more personalized than offline advertising, through the use of machine learning models and real-time auctions for ad targeting. One specific task,  predicting the likelihood of conversion (i.e.\ the probability a user will purchase the advertised product), is crucial to the advertising ecosystem for both targeting and pricing ads. Currently, these models are often trained by observing individual user behavior, but, increasingly, regulatory and technical constraints are requiring privacy-preserving approaches. For example, major platforms are moving to restrict tracking individual user events across multiple applications, and governments around the world have shown steadily more interest in regulating the use of personal data. Instead of receiving data about individual user behavior, advertisers may receive privacy-preserving feedback, such as the number of installs of an advertised app that resulted from a group of users. In this paper we outline the recent privacy-related changes in the online advertising ecosystem from a machine learning perspective. We provide an overview of the challenges and constraints when learning conversion models in this setting. We introduce a novel approach for training these models that makes use of post-ranking signals. We show using offline experiments on real world data that it outperforms a model relying on opt-in data alone, and significantly reduces model degradation when no individual labels are available. Finally, we discuss future directions for research in this evolving area.
\end{abstract}

\section{Introduction}

The key distinction between online advertising and traditional media advertising has been ad targeting \citep{goldfarb2014different}. In traditional media, targeting can often only be done by region or program, whereas online advertising is typically sold in real-time auctions and highly personalized \citep{yuan2013real}. Personalized ad targeting can both improve user experience by showing the user more relevant ads, and provide more value for advertisers \citep{bayer2020impact} and platforms by allocating ad space to the highest value ads for each impression.

Many online ads are \textit{performance ads}, where the advertiser desires the user to take an action such as purchasing an item or installing a mobile application. \textit{Conversion tracking} is used to track which users converted (made a purchase or installed an app) after viewing or clicking the ad and is important to the online advertising ecosystem for two reasons. Firstly, by logging which users converted, personalized models trained on those logs can better predict advertising relevance in future \citep{ma2018entire}. Secondly, by allowing advertisers to determine which ad campaigns and platforms are driving value for them, the problem of advertising attributed to John Wanamaker, ``Half my advertising spend is wasted; the trouble is, I don’t know which half,'' is solved. By being able to measure the performance of their ad campaigns, advertisers can focus ad spending on campaigns that are achieving a profitable cost per conversion.

\begin{figure}
    \centering
    \includegraphics[width=0.7\textwidth]{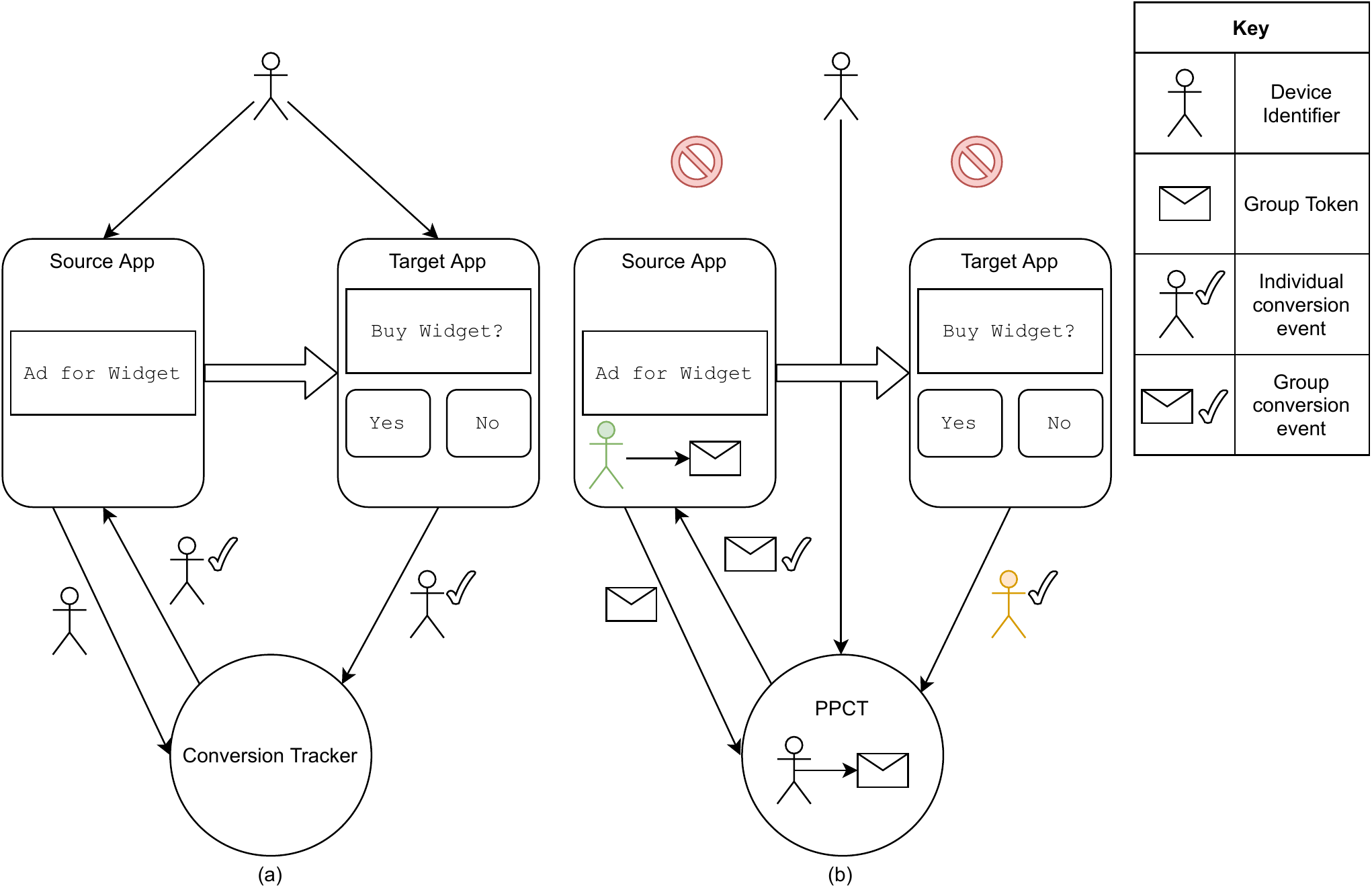}
    \caption{\textbf{(a)}
    The standard approach to conversion tracking on a mobile platform. The user is shown a performance ad in a source app. If the user clicks on the ad it opens the target app where the user can choose to convert (e.g.\ purchase the advertised product). The ad and the conversion take place on different apps so are linked by both apps reporting the action to a conversion tracking service keyed by a unique device identifier. The conversion tracker uses this to link the two actions and provide the individual conversion label to the source app.
    \newline
    \textbf{\textbf{(b)}} In a privacy preserving approach when the user clicks on the ad it opens the target app and passes along a group token. If the user converts then the target app reports the conversion event along with the group token to the source app (typically intermediated by the platform). Crucially, the number of bits conveyed by the group token is extremely limited ($\approx5$ bits) and the reporting is delayed by a varying length of time. These limitations are designed to make it hard to link the conversion event with a specific user while providing feedback on the number of conversions that took place within each group. Both the source and target app may have their own unique user identifiers (indicated by the two different colours), but they are not linked.
    }
    \label{fig:conversiontracking}
\end{figure}

However, conversion tracking has required linking a user's behavior on one service, where they interact with an ad, to their actions on another service (such as a shopping app). This can be done by using a common identifier across apps and reporting relevant user actions linked to this identifier to a third party. Specific examples are Mobile App Conversion Tracking for mobile platforms and third party cookies for web platforms.

Recently, privacy preserving approaches for conversion tracking (PPCT) have been proposed or implemented \citep{willander2019webkit, 2021skad, google2021webconversion, brave2021conversiontracking}.
This is driven by user expectations, regulatory changes, and platform owners placing restrictions for services provided on their platforms.
Most follow a similar approach; advertisers and advertising platforms, instead of receiving information on the conversion of a particular user, now receive delayed feedback about groups of users. 
One timely example is Apple's iOS 14.5 update (released April 2021) which began requiring services to explicitly request user consent in order to access a device identifier for conversion tracking \citep{2021appleprivacy} but also provides a PPCT approach as an alternative \citep{2021skad}.

In this work we provide, to our knowledge, the first overview of the machine learning challenges that these new, privacy preserving paradigms pose to conversion tracking and prediction. We also present a novel approach to solving this problem using post-ranking signals to impute the missing labels. We show in offline experiments on real data that with this approach it is possible to retain $\approx 95\%$ of the conversion prediction accuracy obtained with full label information in a PPCT setting, compared to less than 50\% with a baseline solution. We also show that our method complements opted-in user data, and significantly outperforms the baseline solution at every simulated opt-in rate. Finally, we end with a discussion of future directions for this approach and the industry more generally.

\section{Background} \label{bg}

A large amount of online advertising is performance advertising. Examples include online shopping (the advertiser wants users to open their webstore and make a purchase) \citep{zhu2017optimized} and mobile app installs (the advertiser wants users to install a target app).

In this setup and at large scale, machine learning models are crucial for ad targeting and bidding decisions. Figure \ref{fig:conversiontracking} shows the standard data collection process and the PPCT paradigm. The key challenge is that the ad is shown on one app or website, but the conversion event takes place on a different app or website (e.g. an app store or shopping website). A device identifier or third party cookie is used to link these two events together.

From the ML perspective the logs of these linked events result in a dataset $(x_i, y_i, z_i)$ where $x_i$ is the features available at ranking time describing the user and ad, $y_i \in \{0, 1\}$ is the binary click label indicating if the user clicked on the ad, and $z_i \in \{0, 1\}$ is the binary conversion label indicating if the user converted. Crucially, the conversion label $z_i$ is based on the ability to link the user action of clicking the ad with the user conversion action through the unique identifier. Typically, the conversion label is set to 0 if the user did not click on the ad; $y_i = 0 \rightarrow z_i = 0$.

For the analysis here we assume $z_i$ is a binary label, but it can be broader, for example, the amount spent in an online store. A standard approach is to use the logs to train models which predict click through rate (CTR), $p(y=1|x)$, and conversion rate (CVR), $p(z=1|y=1, x)$. These predictions are used for ad targeting and bidding. For example, if an advertiser is willing to pay $b$ per conversion, then the value of a specific impression $x$ to this advertiser can be estimated as $b \cdot p(z=1|y=1, x) \cdot p(y=1|x)$.

\subsection{Privacy preserving conversion tracking (PPCT)}

Increasingly, the assumptions behind this standard approach to CVR prediction are being challenged by regulatory and platform changes towards stronger user privacy. In particular, it can be desirable or required to avoid explicit linking between user actions on the source app or website, and the target app or website. There are a number of proposed technical approaches to allow for some form of privacy preserving conversion tracking
\citep{2021appleprivacy, willander2019webkit, 2021skad, google2021webconversion, brave2021conversiontracking}.
In most cases these approaches do not provide formal differential privacy guarantees \citep{dwork2014algorithmic} although they may provide k-anonymity \citep{sweeney2002k}. The idea of these approaches is to allow sharing of information about the patterns of groups, while limiting the information that can be inferred about any specific individual.

In PPCT the ad targeting system can still expect to receive $x_i$ and the click label, $y_i$, since these typically all take place on one app or website, but the conversion label, $z_i$, is no longer directly available. Instead, as shown in figure \ref{fig:conversiontracking}, it is replaced by a system that provides noisy group labels. When the ad is clicked, the advertising system can choose a group token, $w_i$, to associate with this click\footnote{This is our generic name for this token, various PPCT systems refer to this as trigger data or campaign id.}. The number of bits allowed for this token is highly limited, typically about 5 bits. On the target app, this conversion event is conveyed back to the advertiser. Crucially, this conversion reporting is both delayed stochastically (typically 24-48 hours) and includes the target app ID and the group token, but does not identify the specific user. This privacy-preserving reporting can be done on-device or through the use of a trusted intermediary such as the platform provider. Additionally, noise may be added or privacy safeguards may also be used, such as limiting reporting of any conversion events for a group token that has had few clicks.

We can transform the noisy callbacks into group labels. That is, we can group all clicks for a target app, with the same group token over some period of time into a group and, based on the conversion reporting, estimate the number of conversions that resulted in this group, $g^z \in \mathbb{Z}^+$. Due to the stochasticity in the reporting time, it can be impossible to be sure that a click was reported within the group, but if the time period is large enough this noise will be small. Thus, the training data would look like $(\{x_{ij}, y_{ij}\}, g_{i}^z)$ where the index $i$ is over the groups, and $j$ the individual examples within the group. Note that the conversion value of any individual in the group is difficult to infer.

One can frame the problem of learning CVR as learning from label proportions \citep{yu2014learning}. The situation here is slightly different, because of the opportunity to actively choose group tokens and thus control how users are grouped. However, because such group tokens may also be used for billing or campaign performance tracking, in practice, the machine learner may have limited control over the way users are grouped, and they may be grouped in a non-random manner which makes learning from label proportions challenging. For example, when testing different ad creatives, cohort grouping may be used to measure the performance, so the mean CVR of the two groups of users may systematically vary.

Because some platforms may continue to allow conversion tracking, and some users may also choose to allow conversion tracking, the training set is likely to contain some examples with individual labels and some examples with only group labels.

Because this is a nascent challenge there is a limited body of work on the specific application challenges. \citep{ayala2021revenue} considered revenue attribution under PPCT, specifically when wishing to predict the lifetime value of the user.

\subsection{Post-ranking signals}

Moving from individual labels to noisy group labels makes conversion (CVR) prediction challenging. One signal we can make use of is post-ranking information. Because CVR prediction is used to determine what ads to show the user we cannot use information that is unknown at that time in the ranking model. However, after the user sees the ad, or clicks on it, we may receive additional post-ranking information, $x'_i$ (e.g.\ the physical coordinates in the ad the user clicked). This information, which some other authors have referred to as privileged information \citep{xu2020privileged}, can't be used as features in the prediction model. However, as we discuss later, it can be used in training the prediction models. All post-ranking signals are obtained from the source app, so no cross-app linking is necessary to obtain these signals.

\subsection{Multitask learning (MTL) aspects}

PPCT will not be adopted universally or instantly. Some platforms or users may adopt it rapidly, while others may choose to allow conversion tracking at an individual level.
This presents an extra challenge, as the fraction of individual labels available may vary. We aim for a solution that performs well across a wide-range of individual label availability.

Because we may be getting individual labels from some platforms and group labels from other platforms, the group labels are not necessarily aggregate labels of the same task. Potentially, the group label examples can be considered to be from a different distribution. This can also be true for users who opt-in to conversion tracking, who may not just be a random subset of all users, but rather behave differently from users who do not opt-in. Additionally, different implementations of PPCT mean that the labels may not be exactly comparable across platforms. There may also be variance by ad or app. For example, users may be more privacy conscious for some types of conversion events. One approach to these challenges is to treat different platforms and user types as distinct tasks in a multitask setup.

A related MTL problem is to use the (less sparse, easier to obtain) click labels to improve user representations for conversion prediction \citep{obrien2021mtl, ma2018entire}. In the absence of conversions labels, this sort of approach is likely to be even more beneficial, and can be combined with the approaches outlined here, but we do not explore it in this work.

\subsection{Fingerprinting and probabilistic matching}

Another approach to recovering signal would be to probabilistically match user events between apps. For example, device fingerprinting \citep{nikiforakis2013cookieless} or other signals such as IP address could be used to estimate which click events resulted in a conversion, thus approximately recovering individual user conversion labels. However, through both technological efforts to counter this approach and regulatory changes, this may not be a sustainable solution or may not meet user expectations. Therefore we do not explore this approach in this work.

\section{Methods}

The key idea of our approach is to use the post-ranking signals to impute a \textit{soft label}, $\hat{z} \in [0, 1]$, for examples where a \textit{hard label}, $z \in \{0, 1\}$, is not available. The training data consists of both individual labeled examples $D_{c} = (x_i, x'_i, y_i, z_i)$ (from opt-in users and platforms with conversion tracking) as well as examples lacking individual labels $D_{ppct} = (x_i, x'_i, y_i)$.

We fit a logistic regression model using only post-ranking signals on the individual labeled examples $w^* = \min_w \sum_{i \in D_{c}} -z_i \log \sigma(x'_i \cdot w) - (1 - z_i) \log (1 - \sigma(x'_i \cdot w))$. Since there are a relatively small number of features in $x'$ we use logistic regression (rather than a more expressive model) to avoid over-fitting. Then, we impute labels for the unlabeled items in $D_{ppct}$, $\hat{z}_i = \sigma(x'_i \cdot w^*)$. Our CVR prediction model, which is a deep neural network, $f_\theta(x_i)$, is then trained using both datasets, $D_{c} \bigcup D_{ppct}$, with a cross-entropy loss and using the imputed soft labels for training examples in $D_{ppct}$. This model cannot use the post-ranking signals as input, since it is required to make predictions before those signals are available.

\subsection{Group labels} \label{group_labels}
Note that our solution does not require group labels. The post-ranking signals alone can be used to construct a soft label. However, the group label information, assuming it is provided, could be used in a number of different ways that would likely improve performance. For instance, (1) the group labels could be an input to the function used to construct soft labels, (2) the group labels could be used to create an additional group loss during model training, or (3) the group labels could be used as part of a calibration stage (this could be done either to the soft labels, or to the predictions generated by the trained conversion model). Using the group labels in this way would be expected to have the most impact if the behavior of users in the PPCT dataset differs from the hard labeled dataset.

\section{Results} \label{results}

In order to measure the performance of our approach we used a dataset where all examples have an individual label, $z_i$, available. Our dataset consists of real-world Android and iOS users, logged in an ad ranking system on Twitter with 10s of millions of rows and a large number of features. We simulated removing varying fractions of iOS users' conversion labels, while retaining conversion tracking for Android users. This allowed us to measure the performance of the model under varying levels of conversion tracking availability, including an upper bound with all individual labels available.

For our experiments here we only used the group labels to calibrate the soft labels. We did this because the data was simulated. That is, conversion tracked and non conversion tracked users were randomly sampled, hence from an identical distribution. In online experiments, where this does not hold, we expect to make use of group labels as discussed in Section \ref{group_labels}.

A lack of individual labels impacts both training but also many regularization techniques, including early stopping which we used in our experiments. We found (data not shown) that when training with few individual labels, models tended to overfit even with early stopping. We found the soft labels method alleviated this issue.

\begin{figure}[ht]
    \centering
    \includegraphics[width=0.6\textwidth]{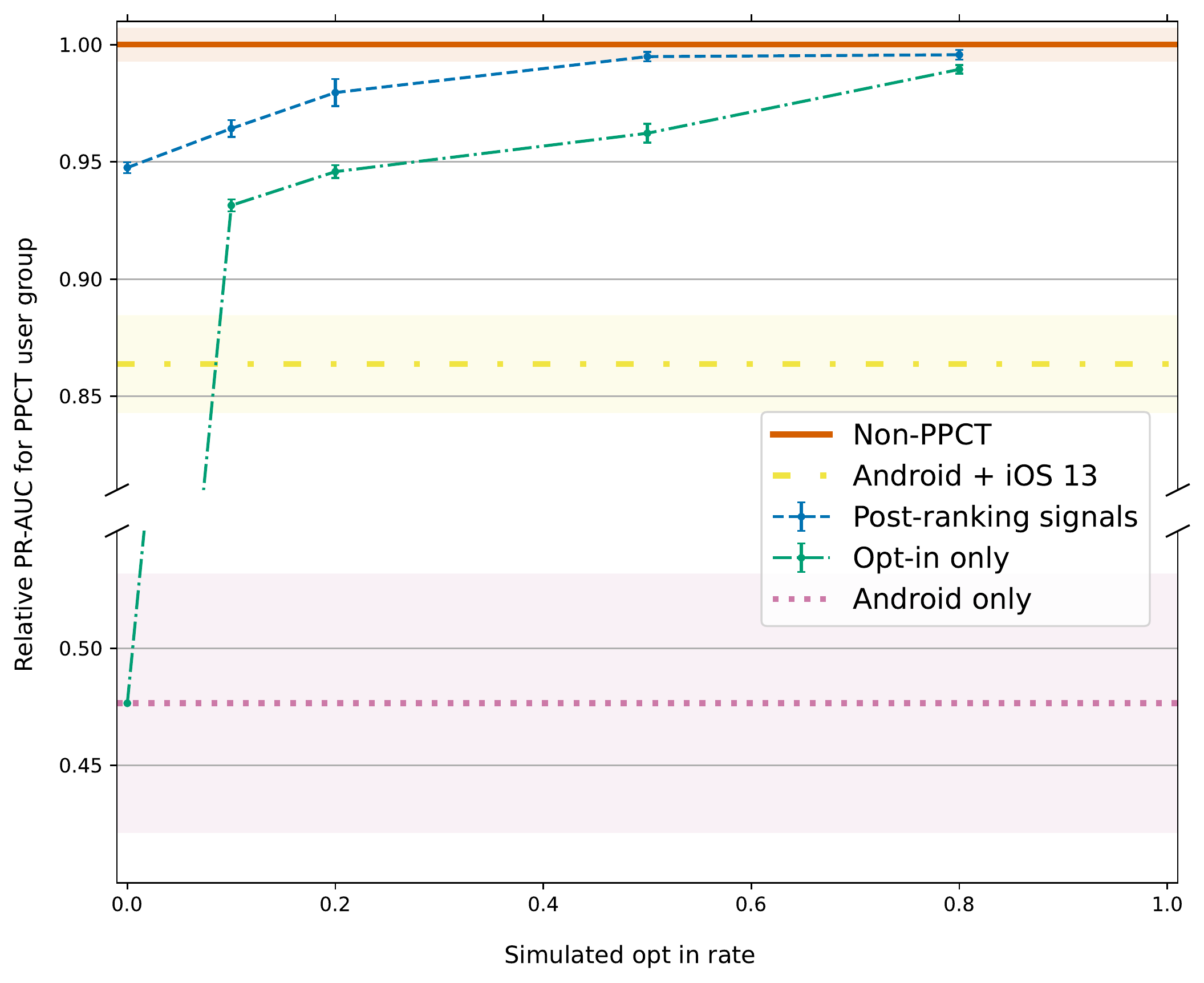}
    \caption{Imputed soft labels dramatically reduce model degradation as traditional hard labels are removed. We use the non-PPCT (all individual labels) as the baseline. ``Android only'', shows that performance degrades to less than 50\% of this baseline if no individual labels are available for iOS users. Our approach, ``Post-ranking signals'', consistently beats relying on opted-in user data only (``Opt-in only''). This difference becomes more significant as the opt-in rate moves to zero, showing how valuable this method could be in the medium/long term scenario where there are few hard labels. When all iOS hard labels are removed, i.e. 0\% opt-in, the soft label model still retains $>94\%$ of the predictive performance of the baseline with all labels available. All errorbars and shading show SE.}
    \label{fig:results_pr_auc}
\end{figure}

\subsection{Experiment settings}

\begin{itemize}
    \item \textbf{Non-PPCT}: This is the learning setting prior to PPCT. We receive all individual conversion labels from Android and iOS. The results for this setting represent an `upper bound' on what could be achieved under a PPCT learning regime.
    \item \textbf{Android only}: We retain only the information for Android users. This comparison point is meant to illustrate a `worst case' when no iOS users choose to allow conversion tracking.
    \item \textbf{Android + iOS $\leq$ 13}: In this setting we retain per-row labels for Android, and iOS users on devices with an OS version less than or equal to 13. In practice, these users represent about 8-9\% of all iOS users in this dataset. iOS users in turn represent about 40-55\% percent of all samples. This setting should simulate the effect of still having labels from late adopters to new technology.
    \item \textbf{Opt-in only}: This setting is identical to ``Android + iOS $\leq$ 13", with the inclusion of some percentage of iOS 14 users.
    \item \textbf{Post-ranking signals}: (our method) This is the same as ``Opt-in only'', except missing labels are imputed with soft labels. All Android and any iOS hard labels available are used, augmented by the soft labels.
\end{itemize}

\subsection{Utility of soft labels}
Figure \ref{fig:results_pr_auc} clearly demonstrates that soft labels are a viable strategy to prevent significant model degradation in a PPCT setting. When 80\% of users are simulated as providing hard labels there is almost no benefit from soft labels. However, as the relative proportion of this data decreases soft labels show significant value, retaining nearly 95\% of the original model performance when all iOS hard labels are lost. This is a substantial improvement over the models that rely on hard labels only, which degrade by over 50\% - this is what would eventually happen to existing systems with the introduction of PPCT if no adaptation is implemented. We also highlight that relying on `late adopters` does not work well. Just like Android, users that do not upgrade (i.e. iOS 13 and below) seem to have conversion behaviour that is distributed differently to users that upgrade, hence the significant drop in performance for ``Android + iOS $\leq$ 13".

Note that ``Android + iOS 13" used a different stopping criteria than the other training settings (which used early stopping) due to its severe overfitting. We learnt an approximate number of epochs that we could train for before overfitting and then only trained for this \textit{fixed} number. We did not do that for ``Opt-in only" because it is unclear how you could know this fixed value in a real world setting as opt-in was constantly changing. Even granting ``Android + iOS 13" this unrealistic advantage is generous. This explains why ``Opt-in only`` can degrade further than ``Android + iOS 13" when at 0\% opt-in they would effectively have the same training data.

Lastly, we expect these results to underestimate the benefit soft labels can bring. In simulation we \textit{randomly} opted users in, but in practice, such users are unlikely to be a random subset of a given group. If they are not random, they likely have different conversion behaviours, and therefore using opted-in user data to learn from will be difficult (as the Android + iOS $\leq$ 13 version examples demonstrate).

\subsection{MTL}

In the results above, we trained a single model, imputing soft labels for missing hard labels. As stated earlier, group labels may be provided from different sources to the hard labels. This combined with the different user distributions just mentioned suggest that the tasks might be better modeled as a multitask learning problem. We have tried an MTL approach using shared parameters and different ``heads'' for soft and hard labels. Our preliminary results (not shown) indicate this approach typically results in better calibration than the single model approach and seems like a promising direction.

\section{Discussion}

As the industry moves towards approaches to targeted advertising that are privacy preserving, both for conversion tracking and in other tasks, we expect the general approach here to be useful. As the information about a specific task become noisier, making use of opt-in data and related task signals becomes crucial. In future, we hope to report on online results of this approach as well as try alternative approaches.

In addition to privacy preserving on-server learning as we have presented here, other privacy preserving approaches may also grow in importance. One important class of approaches we did not make use of here is on-device learning and federated learning. In federated learning the model parameters are updated on device, and these updates communicated back to a central server \citep{kairouz2019advances}, usually with a differential privacy guarantee to ensure that the updates strongly limit what can be inferred about an individual user \citep{geyer2017differentially}. On-device learning refers to approaches where the ad targeting model remains on the user device, and no user-specific information is communicated back to the server. Instead, the server communicates to the device a variety of ad campaigns and budgets, and targeting decisions are made on device \citep[e.g.][]{brave2020ondevice}.

Another approach to PPCT changes that we expect to see happen to some extent in the industry is to avoid the problem by architecting mobile applications so that both the advertising and conversion event take place within a single app. This avoids the need to link conversion outcomes between source and target app. This may have disadvantages such as reducing flexibility by coupling advertising and storefront systems. We did not explore such approaches in this work.

The increasing regulatory and technical constraints on the use of user data also impacts on reproducibility and comparison of results. Even within an organization user data must often be deleted after some time, so the results of an experiment may change each time it is run. Additionally, sharing datasets with outside groups is often very challenging. One potential direction is the use of simulated datasets for research, although that is a large enough topic we did not pursue it here.

\section{Conclusion}

In this work we have outlined, from an ML perspective, the shift from conversion tracking with individual labels to privacy preserving conversion tracking. We have provided a solution to conversion tracking in a PPCT regime by using post-ranking signals and suggested incorporating multitask learning. Tested on an industry-scale dataset we find that without individual conversion labels prediction accuracy drops massively, whereas our soft label approach is capable of attaining only a slightly reduced performance compared to a baseline with all conversion labels available.

\section{Ethical considerations}

The research in the submitted paper has been reviewed as part of our organisation’s research and publishing process. This includes privacy and legal review to help ensure that all necessary obligations are satisfied.

\bibliographystyle{ACM-Reference-Format}
\bibliography{lat}


\end{document}